\documentclass[10pt,twocolumn,letterpaper]{article}

\usepackage{cvpr}              % To produce the CAMERA-READY version
\usepackage{marvosym}
\usepackage{amssymb}
\usepackage{graphicx}
\usepackage{amsmath}
\usepackage{booktabs}
\usepackage{rotating}
\usepackage{multirow}
\usepackage{enumitem}
\usepackage{algorithmicx,algorithm}
\usepackage{enumitem}
\usepackage{pifont}
\usepackage{makeidx}
\usepackage{epstopdf}
\usepackage[noend]{algpseudocode}
\usepackage{xcolor}
\usepackage{colortbl}
\usepackage{color}
\usepackage[pagebackref,breaklinks,colorlinks]{hyperref}
\usepackage[capitalize]{cleveref}
\crefname{section}{Sec.}{Secs.}
\Crefname{section}{Section}{Sections}
\Crefname{table}{Table}{Tables}
\crefname{table}{Tab.}{Tabs.}
\colorlet{dark-blue}{blue!70!black}
\hypersetup{
    colorlinks=true,%
    citecolor=dark-blue,%
    filecolor=dark-blue,%
    linkcolor=red,%
    urlcolor=magenta
}

\usepackage{caption}
\def\cvprPaperID{8460} % *** Enter the CVPR Paper ID here
\def\confName{CVPR}
\def\confYear{2024}

\makeatletter
\def\thanks#1{\protected@xdef\@thanks{\@thanks
        \protect\footnotetext{#1}}}
\makeatother

\begin{document}

\title{Semantic Connectivity-Driven Pseudo-labeling for Cross-domain Segmentation}
%\title{Stable Neighbor Denoising for Source-free Domain Adaptive Semantic Segmentation. }

%\thanks{This work is supported by the National Natural Science Foundation of China(No.62271377, No.62201407),  the Key Research and Development Program of Shannxi (No.2021ZDLGY01-06, No.2022ZDLGY01-12),  the National Key R$\&$D Program of China (No. 2021ZD0110404), the China Postdoctoral Science Foundation (No. 2022M722496), the Foreign Scholars in University Research and Teaching Program’s 111 Project (B07048).}

\author{Dong Zhao$ ^{1}$, Ruizhi Yang$ ^{1}$, Shuang Wang$ ^{1}$ \textsuperscript{\Letter}, Qi Zang$ ^{1}$, Yang Hu$ ^{1}$, Licheng Jiao$ ^{1}$, \\ 
Nicu Sebe$ ^{2}$, Zhun Zhong$ ^{3}$
 \\
$ ^{1}$ School of Artificial Intelligence,
Xidian University, Shaanxi, China \\
$ ^{2}$ Department of Information Engineering and Computer Science, University of Trento, Italy \\
$ ^{3}$ School of Computer Science, University of Nottingham, United Kingdom \\
%\and
%Shuang Wang\\
}
\maketitle

%%%%%%%%% ABSTRACT
\begin{abstract}
%Unsupervised domain adaptation (UDA) addresses the challenge of diminishing adaptability in deep semantic segmentation models when exposed to open domains.
Presently, self-training stands as a prevailing approach in cross-domain semantic segmentation, enhancing model efficacy by training with pixels assigned with reliable pseudo-labels. 
However, we find two critical limitations in this paradigm.
(1) The majority of reliable pixels exhibit a speckle-shaped pattern and are primarily located in the central semantic region. This presents challenges for the model in accurately learning semantics.
(2) Category noise in speckle pixels is difficult to locate and correct, leading to error accumulation in self-training.
To address these limitations, we propose a novel approach called Semantic Connectivity-driven pseudo-labeling (SeCo). This approach formulates pseudo-labels at the connectivity level and thus can facilitate learning structured and low-noise semantics.
Specifically, SeCo comprises two key components: Pixel Semantic Aggregation (PSA) and Semantic Connectivity Correction (SCC). Initially, PSA divides semantics into ``stuff'' and ``things'' categories and aggregates speckled pseudo-labels into semantic connectivity through efficient interaction with the Segment Anything Model (SAM).
This enables us not only to obtain accurate boundaries but also simplifies noise localization. 
%This simplicity arises from the fact that distinguishing noise at the connectivity level is much easier than at the pixel level.
Subsequently, SCC introduces a simple connectivity classification task, which enables locating and correcting connectivity noise with the guidance of loss distribution. Extensive experiments demonstrate that SeCo can be flexibly applied to various cross-domain semantic segmentation tasks, including traditional unsupervised, source-free, and black-box domain adaptation, significantly improving the performance of existing state-of-the-art methods. The code is available at \href{https://github.com/DZhaoXd/SeCo}{https://github.com/DZhaoXd/SeCo}
\end{abstract}

\begin{figure*}[tbp]
    \begin{center}
    \centering \includegraphics[width=0.95\textwidth, height=0.39\textwidth]{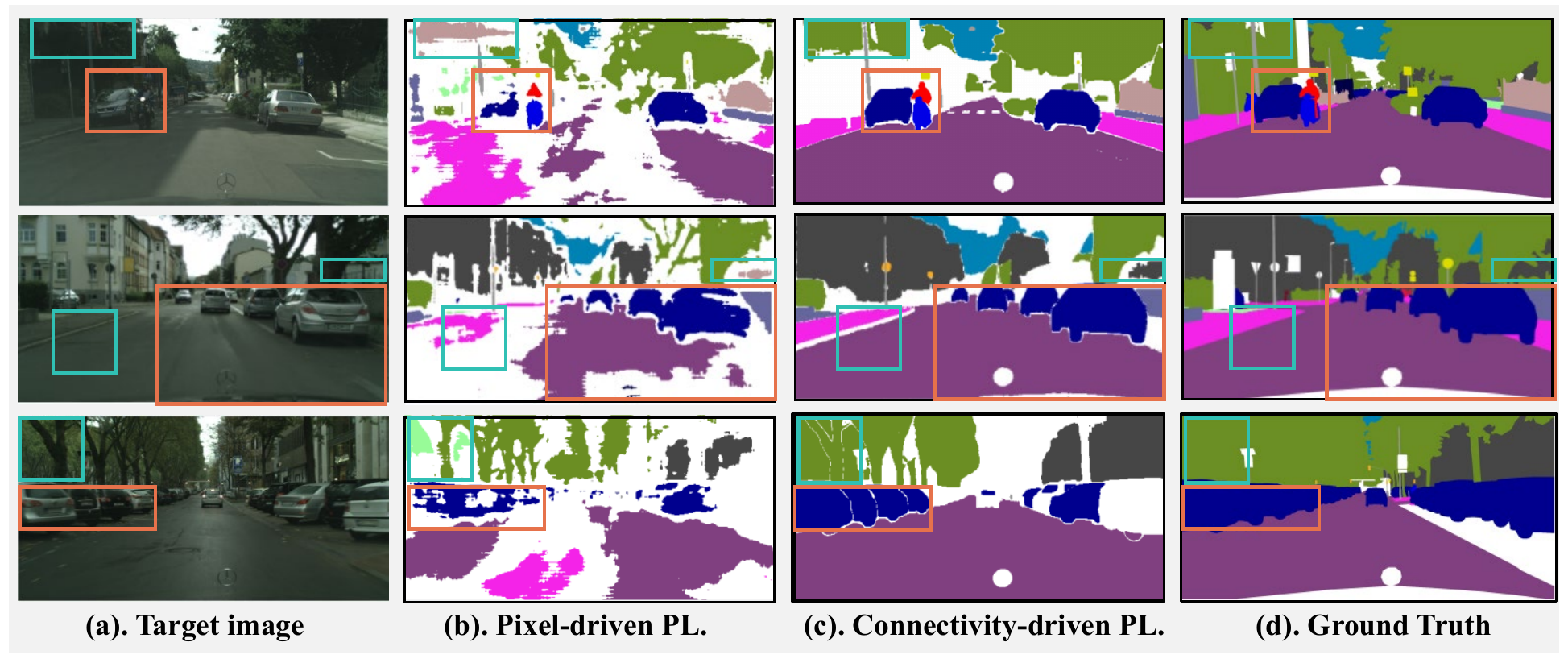}
    \end{center}
    \vspace{-.18in}
%    \captionsetup{font={small}}
    \caption{
Comparison of pixel-driven pseudo-labels (a) and the proposed semantic connectivity-driven pseudo-labels (b). Our approach effectively enhances the pseudo-labels with structured (\textcolor{orange}{the orange box}) and low-noise semantics (\textcolor{cyan}{the cyan box}). White regions in pseudo-label denote filtered areas.
} 
    \label{compPL}
\vspace{-0.05cm}
\end{figure*}

\vspace{-.15in}
\section{Introduction}
\label{sec:intro}
Propelled by deep neural networks, remarkable strides have been achieved in semantic segmentation technology\cite{chen2018encoder, cheng2021per, xie2021segformer, jain2023oneformer}. 
However, deep segmentation models encounter a significant decline in adaptability when confronted with open domains. This challenge is mainly attributed to the inherent domain shift between the training and testing data~\cite{DBLP:journals/corr/HoffmanWYD16, Chen_2017_ICCV, csurka2021unsupervised}.
In response to this challenge, Unsupervised Domain Adaptation (UDA)~\cite{Zhang_2018_CVPR, sakaridis2019guided, choi2019self, pmlr-v80-hoffman18a} has been introduced to enhance the adaptability of segmentation models to the target domain. 

Previous UDA methods focus on maximizing the utilization of source data~\cite{tranheden2021dacs, huang2021fsdr, yue2019domain, zhao2022style}, aligning the distributions between the source and target domains~\cite{Ma_2021_CVPR, 9961139, Tsai_2019_ICCV}, and uncovering distinctive features within the target domain~\cite{2020Unsupervised, zheng2021rectifying, Araslanov_2021_CVPR, zhang2021prototypical, hoyer2022daformer, hoyer2022hrda, hoyer2023mic}. Currently, self-training~\cite{yang2022st++, sohn2020fixmatch} emerges as a particularly effective and promising method for delving into target characteristics. 
It encompasses the selection of reliable pseudo-labels through strategies like thresholding~\cite{Araslanov_2021_CVPR}, model voting~\cite{zheng2021rectifying}, and multi-space consistency~\cite{zheng2021rectifying, hoyer2022daformer}, followed by retraining the model to gradually improve its adaptability to the target domain.

Nevertheless, despite the significant advancements made by these methods, we have identified limitations in the pseudo-labels generated by the self-training paradigm, as depicted in Fig.~\ref{compPL} (a). Firstly, the majority of filtered reliable pixels exhibit a speckle-shaped pattern, primarily concentrated at the center of semantic regions (\textcolor{orange}{the orange box}).
This poses a formidable challenge for precise semantic learning, as segmentation edges play a crucial role in learning structured representations and semantic understanding\cite{huang2019ccnet, yu2020context}.
%presenting a formidable challenge for precise semantic learning. This challenge arises due to the critical importance of segmentation edges in learning structured representations and semantic understanding\cite{huang2019ccnet, yu2020context}.
Secondly, category noise within speckle pixels is difficult to pinpoint and correct (\textcolor{cyan}{the cyan box}), as it necessitates the integration of global semantics to determine specific pixel categories. Consequently, training with labels containing speckle-type noise results in error accumulation during subsequent self-training periods.

To overcome these challenges, we propose a novel method named Semantic Connectivity-driven pseudo-labeling (SeCo), which formulates pseudo-labels at the connectivity level for obtaining structured and low-noise semantics. SeCo comprises two key components: Pixel Semantic Aggregation (PSA) and Semantic Connectivity Correction (SCC). Initially, PSA splits the categories into the ``stuff'' and ``things'' forms. Then, PSA efficiently aggregates speckled pseudo-labels into semantic connectivity by interacting with the Segment Anything Model (SAM) \cite{kirillov2023segany}.
%PSA efficiently aggregates speckled pseudo-labels into semantic connectivity by interacting with the Segment Anything Model (SAM) \cite{kirillov2023segany} in the form of stuff and things.
%
This strategy not only ensures precise boundaries but also streamlines noise localization, as distinguishing noise at the connectivity level is inherently more straightforward than at the pixel level. 
Subsequently, SCC introduces a sample connectivity classification task for learning noisy labels within the connectivity. 
As connectivity classification focuses on local overall categories, we propose to leverage the technique of \emph{early learning} in noisy label learning~\cite{zhang2021understanding,liu2020early} to identify connectivity noise, guided by loss distribution.
%
%we can leverage the technique of \emph{early learning} in noisy label learning~\cite{zhang2021understanding,liu2020early}, a utilizing loss distribution to pinpoint connectivity noise.
As illustrated in Fig.~\ref{compPL}(b), the incorporation of the proposed connectivity-driven pseudo-labels significantly enhances the quality of pseudo-labels, showcasing complete structures and reduced category noise.

In summary, the contributions of this paper are threefold.  
First, we identify the drawbacks of existing self-training approaches and highlight the significance of semantic connectivity in addressing these challenges. 
Second, we propose a Semantic Connectivity-driven pseudo-labeling (SeCo) algorithm, which can effectively generate high-quality pseudo-labels, thereby facilitating robust domain adaptation. 
Third, extensive experiments underscore the versatility of SeCo in effectively addressing various cross-domain semantic segmentation tasks, including conventional unsupervised, source-free\cite{kundu2021generalize, huang2021model}, and black-box\cite{Zhang_2023_ICCV} domain adaptation. Notably, SeCo achieves marked enhancements in the more challenging source-free and black-box domain adaptation tasks.

% For instance, in the GTA $\rightarrow$ Cityscape scenario, it increases the mIoU scores of the baseline by $7.1\%$ and $10.4\%$ on source-free and black-box domain adaptation, respectively.

% Extensive experimentation underscores the versatility of SeCo in effectively addressing various cross-domain semantic segmentation tasks, including conventional unsupervised, source-free\cite{kundu2021generalize, huang2021model}, and black-box\cite{Zhang_2023_ICCV} domain adaptation. Notably, SeCo achieves even greater improvements in more challenging source-free and black-box domain adaptation tasks. For instance, in the GTA $\rightarrow$ Cityscape scenario, it elevates the mIoU scores by $7.1\%$ and $10.4\%$ compared to the baseline methods on source-free and black-box domain adaptation, respectively.

%\begin{figure}[tbp]
%    \begin{center}
%    \centering 
%    \includegraphics[width=0.5\textwidth, height=0.225\textwidth]{intro.pdf}
%    \end{center}
%%    \setlength{\abovecaptionskip}{10pt} 
%    \setlength{\abovecaptionskip}{-0.1 cm}
%%    \captionsetup{font={small}}
%    \caption{Comparison of original pixel-driven pseudo-labels (b) and ours semantic connectivity-driven pseudo-labels (c). Our method refines the original pseudo-label with structured ({light blue box}) and low-noise semantics (light yellow box).
%} 
%    \label{advantage}
%\vspace{-0.45cm}
%\end{figure}

\section{Related Work}

\noindent \textbf{Unsupervised Domain Adaptation} (UDA) transfer the source knowledge to target mainly through the following avenues:
\emph{Source Domain Augmentation}: This approach involves employing style augmentation \cite{Yang_2020_CVPR, huang2021fsdr, zhao2022style, lee2022wildnet} and domain randomization \cite{gong2021cluster, yue2019domain, kim2020learning, huang2021rda, he2023patch} to expand the representation space learned by the source domain model with limited data, thereby enhancing the model's generalization capability.
\emph{Minority Class Enhancement}: This line of work introduces minority class resampling\cite{tranheden2021dacs, gao2021dsp, zhao2023towards, ma2023datacentric}, minority class perturbation\cite{wang2023balancing, Ma_2023_CVPR}, and minority class feature alignment \cite{ma2023delving} to enhance the adaptation capability of minority classes.
\emph{Aligning Source and Target Domains:} This line of work employ various domain alignment strategies, \emph{e.g.}, adversarial training \cite{Hong_2018_CVPR, FADA_ECCV_2020}, statistical matching\cite{SIM_2020_CVPR}, across diverse alignment spaces (\emph{e.g.}, input\cite{pmlr-v80-hoffman18a, vu2019advent}, feature\cite{FADA_ECCV_2020} and output space\cite{Tsai_2018_CVPR}) to reduce statistical differences between the two domains.
\emph{Self-Training Techniques}: This line of methods primarily employs pseudo-labeling techniques to further address the issue of inadequate target adaptation. To counter pseudo-label noise, existing approaches employ various strategies, including introducing strong augmentations from input data\cite{hoyer2023mic}, designing teacher-student model structures\cite{hoyer2022hrda}, and employing pseudo-label selection methods\cite{araslanov2021self, li2022class, Pan_2020_CVPR, zou2018unsupervised, Zou_2019_ICCV, mei2020instance, Zhao_2023_ICCV, 10181233}, aiming to alleviate the issue of error accumulation.

\begin{figure*}[tbp]
    \begin{center}
    \centering 
    \includegraphics[width=0.9\textwidth, height=0.425\textwidth]{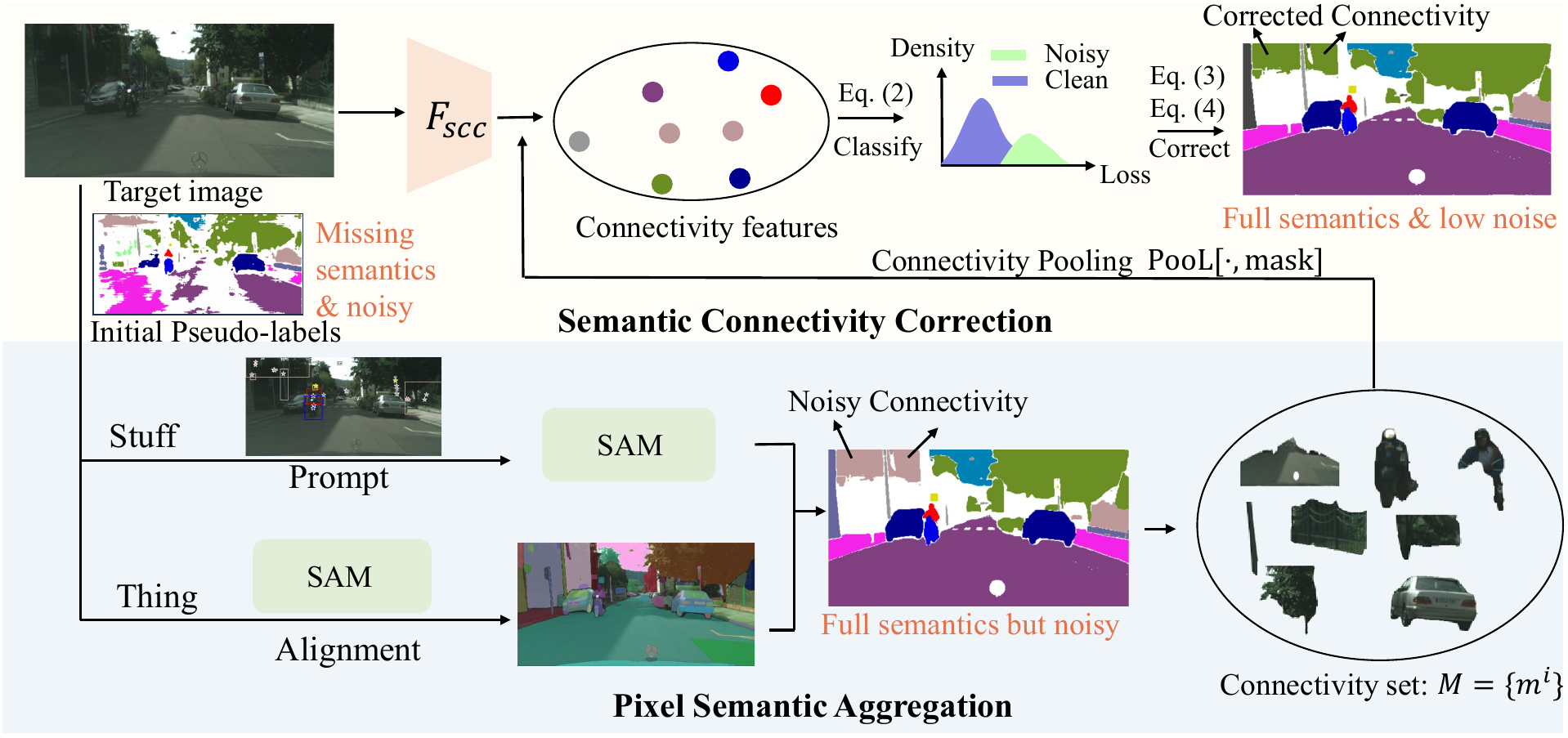}
    \end{center}
%    \setlength{\abovecaptionskip}{10pt} 
    % \setlength{\abovecaptionskip}{-0.1 cm}
%    \captionsetup{font={small}}
    \vspace{-.15in}
    \caption{The framework of the proposed semantic connectivity-driven self-training (SeCo) approach. } 
    \label{pipeline}
\vspace{-0.45cm}
\end{figure*}

\noindent \textbf{Segment Anything Model} (SAM) \cite{kirillov2023segany} has gained widespread attention, with multiple works incorporating it into specific segmentation tasks. For instance, \cite{ma2023segment} fine-tuned SAM in the medical domain to establish a robust foundational medical model. In few-shot learning, \cite{zhang2023personalize} applied SAM, achieving notable results with minimal parameter fine-tuning. \cite{wu2023medical} proposed an efficient method for fine-tuning SAM in downstream segmentation scenarios. Moreover, \cite{chen2023semantic} combined SAM with semantic segmentation models to enhance segmentation model boundaries. Our approach marks the first use of SAM in pseudo-label-based cross-domain segmentation tasks. While \cite{chen2023semantic} is closely related, we conduct a detailed analysis of its limitations in pseudo-label learning, encompassing both methodological and experimental aspects.

\section{Method}
\label{sec:method}

\noindent\textbf{Problem Definition}. Cross-domain semantic segmentation aims to transfer a segmentor trained on the labeled source domain $ \ {D}_{s}=\{(x^{i}_{s}, y^{i}_{s})\}_{i=1}^{i=N_{s}} $ to the unlabeled target domain  $ \ {D}_{t}=\{(x^{i}_{t})\}_{i=1}^{i=N_{t}} $, where $N_{s}$ and $N_{t}$ indicate the number of samples for each domain respectively. $x$ and $y$ represent an image and corresponding ground-truth label, respectively.
Presently, mainstream cross-domain segmentation methods optimize the following objective to enhance model adaptability,
\begin{equation} \label{base_certain}
\mathcal L = \mathcal L_s(x_{s},  y_{s}) + \beta \mathcal L_t(x_{t}, \hat y_{t}),
\end{equation}
where $\mathcal L_s$ is the supervised cross-entropy loss, $\beta$ is the trade-off weight, $\mathcal L_t$ is the unsupervised pseudo-labeling loss, and $\hat y_{t}$ is the pseudo-label.
%It is worth noting that, in the stricter cross-domain tasks, \emph{e.g.}, source-free and black-box domain adaptation where the source data is not accessible, only $\mathcal L_t$ can be applied. 
This formula underscores the critical importance of the quality of pseudo-labels in improving the model's cross-domain ability.
To alleviate the noise in pseudo-labels, various estimation \cite{zheng2021rectifying, hoyer2022daformer, hoyer2022hrda, araslanov2021self, fleuret2021uncertainty} and calibration \cite{wang2022continual, chen2022deliberated, li2022class, xie2023sepico} methods have been introduced for pseudo-label selection. 
However, as mentioned in Sec. \ref{sec:intro}, the filtered pseudo-labels still encounter issues of constrained semantics and challenging localization of category noise. We next present a novel approach to address these issues from the perspective of semantic connectivity.

\subsection{Overview}
The presented \textbf{Se}mantic \textbf{Co}nnectivity-driven pseudo-labeling (SeCo) is illustrated in Fig.~\ref{pipeline}. SeCo comprises two components, namely Pixel Semantic Aggregation (PSA) and Semantic Connectivity Correction (SCC), working collaboratively to refine the low-quality and noisy pseudo-labels into high-quality and clean pseudo-labels.
Initially,  PSA aggregates pixels from the filtered pseudo-labels into connections by interacting with the \emph{segment anything model} (SAM) \cite{kirillov2023segany} through stuff and things interactions. Subsequently, PSA segments the image into multiple connectivities based on their semantics. Guided by the connectivity set, SCC establishes a connectivity classifier, conducts connectivity pooling on image features, and classifies each connectivity. Leveraging information about fitting difficulty and loss distribution, SCC identifies and corrects noise. 
Finally, the connectivity-driven pseudo-labels, characterized by comprehensive semantics and low noise are achieved.

\subsection{Pixel Semantic Aggregation}

\noindent\textbf{Motivation:~Why SAM?} Pixel semantic aggregation (PSA) proposes utilizing reliable pixels within pseudo-labels as category references and subsequently aggregating pixels that share similar semantics into connections.
Intuitively, the above goals can be achieved through interactive segmentation \cite{sofiiuk2020f, lin2020interactive, ramadan2020survey} or pixel clustering \cite{obukhov2019gated, abdal2021labels4free}, but traditional techniques often struggle to accurately identify semantic boundaries in complex scenes, resulting in ambiguous aggregation.
The advent interactive segmentation model, \emph{segment anything model} (SAM)\cite{kirillov2023segany}, provides powerful semantic capture capabilities. 
With reasonable prompts, SAM has the potential to give accurate semantic boundaries even in complex scenes \cite{ma2023segment, jing2023segment}.
Building on SAM's remarkable capability, we are motivated to investigate how to leverage the reliable yet limited pseudo-labels to prompt SAM effectively and enhance the completion of pseudo-label semantics.

\begin{figure}[tbp]
    \begin{center}
    \centering 
    \includegraphics[width=0.485\textwidth, height=0.35\textwidth]{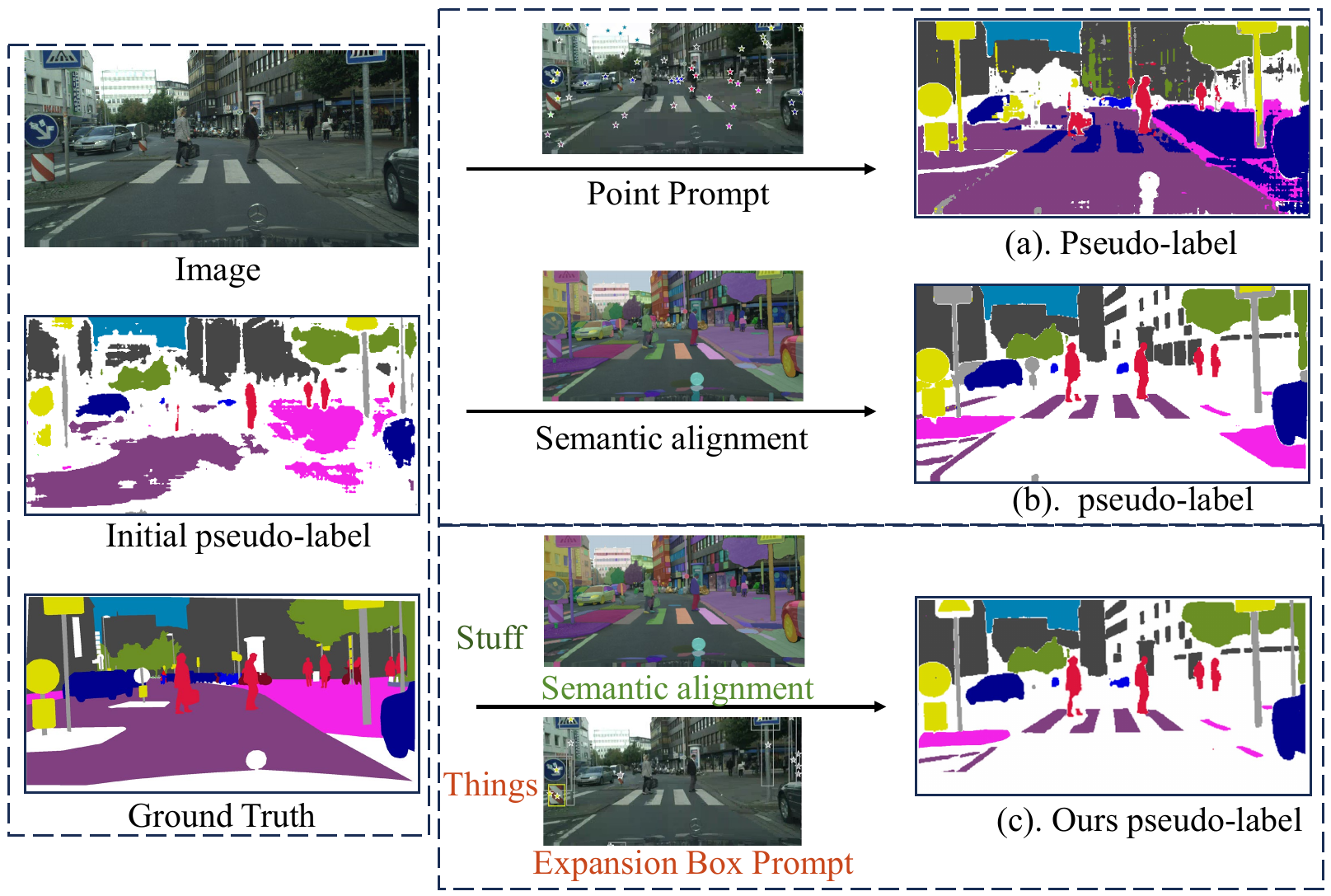}
    \end{center}
%    \setlength{\abovecaptionskip}{10pt} 
    % \setlength{\abovecaptionskip}{-0.1 cm}
%    \captionsetup{font={small}}
\vspace{-.1in}
    \caption{Comparison of pseudo-label aggregation using different interactive methods with SAM \cite{kirillov2023segany}.
} 
    \label{advantage}
\vspace{-.11in}
\end{figure}

\noindent\textbf{Discussion on Utilization of SAM.} We analyze two straightforward solutions as outlined below. The first method involves sampling the center pixels of each connected region on the pseudo-label as prompt points, as depicted in Fig.~\ref{advantage}(a). We observe that when prompt points of the same category contain noise, this method compromises the aggregated segmentation structure. The disruption is attributed to noisy prompts interfering with the cross-attention mechanism in SAM~\cite{kirillov2023segany}. 

The second method represents an improved way, called semantic alignment~\cite{chen2023semantic},  aligning pseudo-labels with the connectivity established by SAM. This involves selecting the pseudo-label with the maximum proportion in each connectivity as the category for the entire region, as illustrated in Fig.~\ref{advantage} (b). We note that while this approach can refine pseudo-labels, it is consistently influenced by SAM's uncertain semantic granularity, particularly in the context of neighboring instance objects.
Fig.~\ref{advantage} (b) provides examples of failures in this method, where SAM aggregates two categories, ``traffic sign'' and ``pole'' into a semantic connected region, leading to misaligned pseudo-labels due to this uncertainty. Our analysis indicates that this issue arises because SAM constructs connectivity by uniformly sampling points in space as prompts and subsequently filtering out redundantly connected regions. This fails to ensure corresponding sampling points for neighboring instance objects, resulting in a semantic granularity deviation between SAM's connectivity and specific segmentation tasks. 

In summary, ``prompts'' interaction can aid in determining semantic granularity but is vulnerable to noise interference; Conversely, ``alignment'' interaction can alleviate noise interference but is susceptible to uncertain semantic granularity.

\noindent\textbf{The Proposed Strategy.} Building upon the analysis above, we find that noise significantly affects ``stuff'' categories due to their larger size and higher pixel proportions, making them more prone to selecting noisy pixels. On the other hand, semantic granularity uncertainty is more prevalent in ``things'' categories, given their smaller size and dense adjacency.
To this end, we propose to interact with SAM in the form of ``stuff'' and ``thing''. Specifically, for stuff, we employ box and point prompts to guide the semantic precision,  while for things,  we utilize semantic alignment to mitigate the impact of noisy prompts.
The detailed algorithm is in Algorithm~\ref{InSAM}. An illustration of the proposed strategy is shown in Fig.~\ref{advantage} (c).

%\cite{chen2023semantic}

\begin{algorithm}[tbp]
  \caption{Aggregation of Pseudo-Labels with SAM}
	\label{InSAM}
  \begin{algorithmic}[1]
    \Procedure{AggregatePseudoLabels}{Image $x$, Pseudo-Label $\hat{y}$, SAM Model}
      \State \textbf{Aggregate Pseudo-Labels for ``things'' category:}
      \State Extract connectivities of $\hat{y}$
      \For{\textbf{each connectivity} in $\hat{y}$ of ``things'' category}
        \State Compute enlarged maximum bounding box as box prompt
        \State Compute geometric center as point prompt 
        \State Interact with SAM using box and point prompts 
        \State Obtain aggregated connectivity
      \EndFor
      \State \textbf{Aggregate Pseudo-Labels for ``stuff'' category:}
      \State Input $x$ into SAM to get no-semantic connectivities 
      \For{\textbf{each ``stuff'' category} in $\hat{y}$}
        \State Align $\hat{y}$ with the no-semantic connectivities by assigning the maximum proportion pseudo-label
        \State Obtain aggregated connectivities 
      \EndFor
      \State \textbf{Merge stuff and thing:}
      \State $\hat{y}_{\text{psa}} \gets$ Merge and filter overlapping connectivities
      \State \textbf{Output:} Aggregated Pseudo-Label $\hat{y}_{\text{psa}}$
    \EndProcedure
  \end{algorithmic}
\end{algorithm}

\subsection{Semantic Connectivity Correction}

PSA aggregates both precise and noisy pseudo-labels into connectivities, facilitating the locating and correction of noise. 
This simplicity arises from the fact that distinguishing noise at the connectivity level is much easier than at the pixel level, as it eliminates the necessity to scrutinize local semantics and instead focuses on the overall category. Inspired by this, we propose Semantic Connectivity Correction (SCC), introducing a simple connectivity classification task and detecting noise through loss distribution.

%With PSA,  both precise and noisy pseudo-labels are used for establishing connectivity.
%Through the efforts of PSA, both precise and noisy pseudo-labels serve as references for spatially clustering similar pixels and establishing connections. 
%This simplifies the task of detecting noise in pseudo-labels, as the focus shifts from scrutinizing the semantic structure around each pixel to determining the overall category of the connection.
%Moreover, when we transition the denoising target from pixel to connectivity, conclusions regarding noisy learning in classification tasks can similarly be extrapolated, as both are concerned with the overall noise of the target.

Specifically, given the input image \(x_i\), we first obtain the connectivity mask list \(M = \{m^{i,n}\}_{n=1}^{n=N_i}\) and its corresponding connectivity-level pseudo-label \(\hat y_{sc} = \{\hat y_{sc}^{i,n}\}_{n=1}^{N_i}\) from PSA, where \(N_i\) represents the number of connectivities for the \(i\)-th sample \(x_i\). Then, we set up a connectivity classifier, comprising a feature extractor \(\rm {F_{scc}}\) and a linear layer \(\rm{MLP}\), and optimize it with the following objective,
\begin{equation}  \label{lscc}
L_{scc} =  \sum_{i, k, n} - \hat y_{sc}^{i,n,k} \log  ({\rm MLP}( {\rm Pool}[{\rm F_{scc}}(x^i), m^{i,n}] )) .
\end{equation}
 \(\rm Pool[\cdot, \text{mask}]\) denotes the average pooling of features corresponding to the input mask, \(k \in [0, 1,... K]\), and \(K\) is the category number. Optimizing \(L_{scc}\) conducts a \(K\)-way classification for each connectivity with clean and noisy labels.

\begin{figure}[tbp]
    \begin{center}
    \centering 
    \includegraphics[width=0.405\textwidth, height=0.30\textwidth]{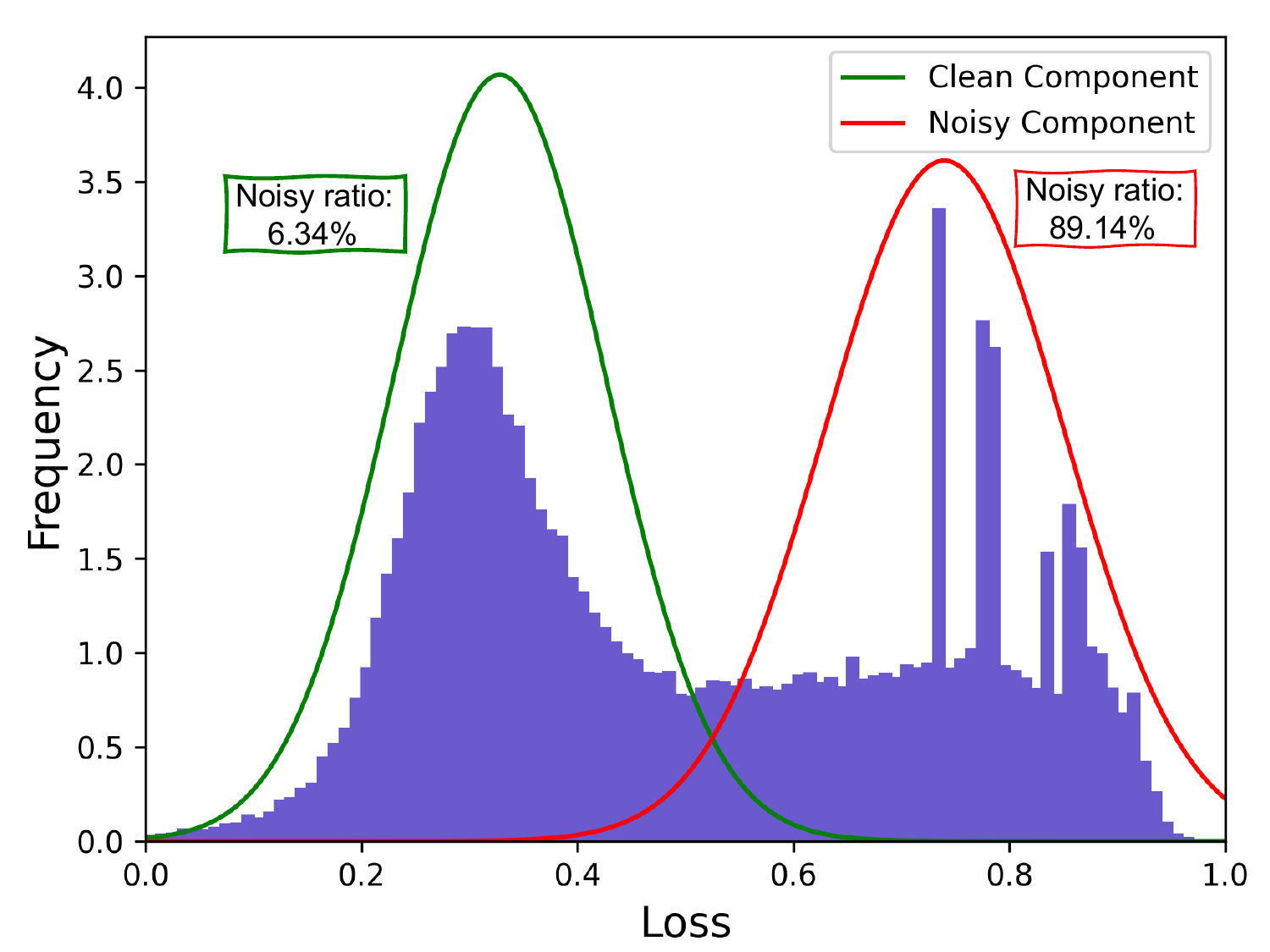}
    \end{center}
    \setlength{\abovecaptionskip}{-0.1 cm}
%    \captionsetup{font={small}}
    \caption{The loss distribution on GTA5 $\rightarrow$ Cityscapes tasks.
} 
    \label{bimodel}
\vspace{-0.45cm}
\end{figure}

Based on observations of \emph{early learning} in noisy learning\cite{zhang2021understanding,liu2020early, yang2022divide}: when training on noisy labels, deep neural networks, in the early stages of learning, initially match the training data with clean labels, and subsequently memorize examples with erroneous labels.
We warm up the connectivity classifier for several epochs and then obtain the loss distribution by calculating Eq.~\ref{lscc} the for each connectivity.
As shown in Fig.~\ref{bimodel}, the loss of connectivities persent bimodol distribution, and the clusters with larger losses correspond to higher noise, which better conform to the observations.
To this end,  we employ a two-component Gaussian Mixture Model to effectively model the loss distribution using the Expectation-Maximization algorithm \cite{yang2022divide}. Subsequently, the probability of a connectivity being noisy, denoted as $\eta$, can be reasonably approximated by the Gaussian distribution associated with bigger loss, \emph{i.e.}, $\eta^{i, n} = p(c |L_{scc}(x, m^{i, n}))$. $c$ is the parameters of the corresponding Gaussian distribution.
We keep the clean connectivity by setting a noise threthod $\tau_{ns}$, \emph{i.e.}, 
\begin{equation}
D_{clean} = \{(x_i,  y_{sc}^{i,n}) | \eta^{i, n} < \tau_{ns} \}.
\end{equation}
Besides, we find that many noisy connectivities can be corrected by setting another correction threshold $\tau_{cr}$ on the connectivity classifier's output probability, \emph{i.e.}, 
\begin{equation}
D_{corr} = \{(x^i,  k) | p^{i, n,k}_{scc} > \tau_{cr}, \eta^{i, n} > \tau_{ns} \}.
\end{equation}
where $p^{n,k}_{scc}$ represents the probability of class $k$ for the $n$-th connectivity.
We take the union of the two sets as the final connectivity-driven pseudo-label set $D_{all} = D_{clean} \cup D_{corr}$ for self-training.

\subsection{Implementation on Domain Adaptation Tasks}
We provide solutions on how to use connectivity-driven pseudo-labels for different domain adaptation tasks.

\noindent \textbf{Unsupervised Domain Adaptation.} 
The connectivity-driven pseudo-label set $D_{all}$ serve two primary functions.
Firstly, they contribute to the pseudo-labeling $L_t$ loss in Eq.~\ref{base_certain}, providing accurate semantic guidance for the target domain. 
The second objective is to mitigate category bias in domain adaptation. 
We treat $D_{all}$ as a sample pool, where we resample minority classes in the target domain and duplicate them through copy-paste operation \cite{gao2021dsp} onto both the source and target datasets.

\noindent \textbf{Source-free \& Black-box Domain Adaptation.}
In these scenarios, source access is restricted. This limitation prevents the deployment of the source loss $L_s$ in Eq.~\ref{base_certain}, making self-training more vulnerable to noise interference. 
Connectivity-driven pseudo-labels $D_{all}$ brings a novel idea to mitigate these challenges.  
With its contribution to accurate semantics and low noise, $D_{all}$ can be viewed as a well-organized labeled set, thereby transforming source-free and black-box domain adaptation tasks into semi-supervised segmentation tasks\cite{st++, unimatch}.

\begin{table*}[tbp]
  \centering
\resizebox{\textwidth}{!}{
    \begin{tabular}{c|cccccccccccccccccccc}
    \toprule
          & \begin{sideways}Road\end{sideways} & \begin{sideways}S.walk\end{sideways} & \begin{sideways}Build.\end{sideways} & \begin{sideways}Wall\end{sideways} & \begin{sideways}Fence\end{sideways} & \begin{sideways}Pole\end{sideways} & \begin{sideways}Tr.Light\end{sideways} & \begin{sideways}sign\end{sideways} & \begin{sideways}Veget.\end{sideways} & \begin{sideways}terrain\end{sideways} & \begin{sideways}Sky\end{sideways} & \begin{sideways}Person\end{sideways} & \begin{sideways}Rider\end{sideways} & \begin{sideways}Car\end{sideways} & \begin{sideways}Truck\end{sideways} & \begin{sideways}Bus\end{sideways} & \begin{sideways}Train\end{sideways} & \begin{sideways}M.bike\end{sideways} & \multicolumn{1}{c|}{\begin{sideways}Bike\end{sideways}} & mIoU \\
    \midrule
    \multicolumn{21}{c}{\textbf{Unspervised domain adaptation: GTA $\rightarrow$ Cityscapes}} \\
    \midrule 
    AdvEnt \cite{ADVENT_2019_CVPR} $\rm ^{ICCV'19}$ & 89.4  & 33.1  & 81.0  & 26.6  & 26.8  & 27.2  & 33.5  & 24.7  & 83.9  & 36.7  & 78.8  & 58.7  & 30.5  & 84.8  & 38.5  & 44.5  & 1.7   & 31.6  & \multicolumn{1}{c|}{32.4 } & 45.5  \\  \rowcolor{gray!15}
    AdvEnt + Ours & 92.0  & 61.0  & 87.0  & 51.0  & 49.4  & 48.9  & 44.5  & 44.3  & 86.7  & 50.0  & 87.9  & 63.3  & 46.0  & 89.7  & 57.6  & 54.6  & 5.6   & 47.7  & \multicolumn{1}{c|}{51.6 } & \textbf{58.9}\rm \textcolor[RGB]{0,155,0}{(+13.4)} \\
    \midrule
    ProDA \cite{zhang2021prototypical} $\rm ^{CVPR'21}$ & 91.5  & 52.4  & 82.9  & 42.0  & 35.7  & 40.0  & 44.4  & 43.3  & 87.0  & 43.8  & 79.5  & 66.5  & 31.4  & 86.7  & 41.1  & 52.5  & 0.0   & 45.4  &  \multicolumn{1}{c|}{53.8} & 53.7  \\ \rowcolor{gray!15}
	 ProDA + Ours & 94.4  & 65.6  & 87.8  & 55.8  & 54.7  & 56.8  & 58.6  & 60.3  & 90.2  & 51.5  & 93.7  & 72.7  & 48.0  & 88.1  & 51.3  & 65.3  & 1.5   &  60.3 &  \multicolumn{1}{c|}{61.0} & \textbf{64.1} \rm \textcolor[RGB]{0,155,0}{(+9.4)}\\
    \midrule
    DAFormer \cite{hoyer2022daformer} $\rm ^{CVPR'22}$ & 95.7  & 70.2  & 89.4  & 53.5  & 48.1  & 49.6  & 55.8  & 59.4  & 89.9  & 47.9  & 92.5  & 72.2  & 44.7  & 92.3  & 74.5  & 78.2  & 65.1  & 55.9  & \multicolumn{1}{c|}{61.8} & 68.2  \\ \rowcolor{gray!15}
	 DAFormer+Ours & 96.2  & 74.4  & 90.9  & 56.7  & 49.7  & 60.5  & 62.7  & 69.4  & 92.4  & 54.9  & 93.9  & 77.1  & 53.1  & 96.6  & 83.1  & 82.2  & 72.5  & 62.6  & \multicolumn{1}{c|}{65.6}   & \textbf{73.4} \rm \textcolor[RGB]{0,155,0}{(+5.3)} \\
    \midrule
    HRDA \cite{hoyer2022hrda} $\rm ^{ECCV'22}$ & 96.4  & 74.4  & 91.0  & 61.6  & 51.5  & 57.1  & 63.9  & 69.3  & 91.3  & 48.4  & 94.2  & 79.0  & 52.9  & 93.9  & 84.1  & 85.7  & 75.9  & 63.9  & \multicolumn{1}{c|}{67.5} & 73.8  \\ \rowcolor{gray!15}
    HRDA+Ours & 96.6  & 80.9  & 92.4  & 62.5  & 57.5  & 61.0  & 66.7  & 71.7  & 92.4  & 52.3  & 95.1  & 80.6  & 56.3  & 95.9  & 86.1  & 86.6  & 76.8  & 65.4  & \multicolumn{1}{c|}{68.7} & \textbf{76.1} \rm \textcolor[RGB]{0,155,0}{(+2.3)} \\
    \midrule
    \multicolumn{21}{c}{\textbf{Source-free domain adaptation:  GTA $\rightarrow$ Cityscapes }} \\
    \midrule
    HCL \cite{huang2021model} $\rm ^{NIPS'21}$  & 92.0  & 55.0  & 80.4  & 33.5  & 24.6  & 37.1  & 35.1  & 28.8  & 83.0  & 37.6  & 82.3  & 59.4  & 27.6  & 83.6  & 32.3  & 36.6  & 14.1  & 28.7  & \multicolumn{1}{c|}{43.0} & 48.1  \\ \rowcolor{gray!15}
    HCL+Ours & 94.6  & 62.5  & 88.6  & 48.4  & 41.6  & 45.2  & 43.5  & 32.9  & 84.0  & 45.3  & 91.6  & 66.0  & 47.5  & 89.0  & 42.6  & 58.8  & 31.5  & 47.2  & \multicolumn{1}{c|}{56.2} & \textbf{58.8} \rm \textcolor[RGB]{0,155,0}{(+10.6)} \\
    \midrule
    DTST \cite{zhao2023towards} $\rm ^{CVPR'23}$  & 90.3  & 47.8  & 84.3  & 38.8  & 22.7  & 32.4  & 41.8  & 41.2  & 85.8  & 42.5  & 87.8  & 62.6  & 37.0  & 82.5  & 25.8  & 32.0  & 29.8  & 48.0  & \multicolumn{1}{c|}{56.9} & 52.1  \\ \rowcolor{gray!15}
    DTST+Ours & 94.9  & 65.9  & 89.9  & 48.2  & 42.3  & 45.9  & 48.9  & 45.6  & 85.7  & 46.2  & 91.1  & 68.2  & 47.6  & 88.5  & 44.9  & 57.8  & 29.5  & 50.7  & \multicolumn{1}{c|}{57.8} & \textbf{60.5} \rm \textcolor[RGB]{0,155,0}{(+8.4)} \\
    \midrule
    \multicolumn{21}{c}{\textbf{Black-box domain adaptation:  GTA $\rightarrow$ Cityscapes }} \\
    \midrule
    DINE \cite{liang2022dine} $\rm ^{CVPR'22}$ & 88.2  & 44.2  & 83.5  & 14.1  & 32.4  & 23.5  & 24.6  & 36.8  & 85.4  & 38.3  & 85.3  & 59.8  & 27.4  & 84.7  & 30.1  & 42.2  & 0.0   & 42.7  &  \multicolumn{1}{c|}{45.3} & 46.7  \\ \rowcolor{gray!15}
    DINE+Ours & 89.6  & 60.8  & 84.1  & 46.3  & 38.4  & 44.0  & 41.6  & 32.2  & 82.1  & 41.7  & 86.6  & 63.4  & 44.9  & 83.9  & 41.5  & 58.6  & 0.0   & 40.5  & \multicolumn{1}{c|}{54.1}   & \textbf{54.4} \textcolor[RGB]{0,155,0}{(+7.7)} \\
    \midrule
    BiMem \cite{Zhang_2023_ICCV} $\rm ^{ICCV'23}$ & 94.2  & 59.5  & 81.7  & 35.2  & 22.9  & 21.6  & 10.0  & 34.3  & 85.2  & 42.4  & 85.0  & 56.8  & 26.4  & 85.6  & 37.2  & 47.4  & 0.2   & 39.9  & \multicolumn{1}{c|}{50.9}  & 48.2  \\ \rowcolor{gray!15}
    BiMem+Ours & 93.9  & 61.4  & 87.6  & 47.7  & 41.3  & 44.0  & 43.2  & 32.7  & 83.2  & 44.4  & 91.4  & 66.9  & 46.6  & 88.7  & 42.6  & 60.8  & 0.0   & 46.2  & \multicolumn{1}{c|}{55.0}  & \textbf{56.7}\rm \textcolor[RGB]{0,155,0}{(+8.5)}  \\
    \bottomrule
    \end{tabular}%
}
\vspace{-.02in}
\caption{Comparison of mIoU scores (\%) for the segmentation performance between SeCo (Ours) and other state-of-the-art methods across three sets of unsupervised domain adaptation tasks, where GTA serves as the source domain.}
\vspace{-.15in}
  \label{GTA}%
\end{table*}%
%$\rm _{\textcolor[RGB]{0,102,0}{+8.5}}$

\section{Experiments}
\subsection{Setup}
\noindent \textbf{Datasets}: We employ two real datasets (Cityscapes \cite{Cordts_2016_CVPR} and BDD-100k \cite{yu2020bdd100k}) alongside two synthetic datasets (GTA5\cite{gta5_dataset} and SYNTHIA\cite{syhth_dataset}). The Cityscapes dataset comprises 2,975 training images and 500 validation images, all with a resolution of 2048×1024. 
BDD-100k  is a real-world dataset compiled from various locations in the United States. It encompasses a variety of scene images, including those captured under different weather conditions such as rain, snow, and clouds, all with a resolution of 1280×720.
The GTA5 dataset consists of 24,966 images with a resolution of 1914×1052, sharing 19 common categories with Cityscapes. The SYNTHIA dataset encompasses 9,400 images with a resolution of 1280×760, featuring 16 common categories with Cityscapes.

\noindent \textbf{Implementation Details.} 
% Baselines \&
\noindent \emph{Traditional UDA}: We opted for two network architectures: deeplab-v2 \cite{deeplab_v2} with ResNet101 \cite{He_2016_CVPR} and SegFormer \cite{xie2021segformer} with MiT-B5. For deeplab-v2, we chose two classical methods, AdvEnt\cite{ADVENT_2019_CVPR} and ProDA\cite{zhang2021prototypical}, as the baselines. We followed their configurations, including the SGD optimizer and learning rate. For SegFormer, we selected two highly successful UDA methods, DAFormer\cite{hoyer2022daformer} and HRDA\cite{hoyer2022hrda}, as the baselines. 
We adopt the Adam optimizer\cite{kingma2014adam} and the learning rate follows their work.
%The parameterization of the copy-paste operation follows \cite{gao2021dsp}.
\noindent \emph{Source-free UDA}:
We maintain deeplab-v2 as the base network to align with existing works. We chose HCL \cite{huang2021model} and the current SOTA method DTST \cite{zhao2023towards} as baselines. %Initially, we employ their work to extract connectivity-driven pseudo-labels, following which we execute a semi-supervised semantic segmentation method \cite{unimatch} for further self-training.
\noindent \emph{Black-box UDA}:
We use two SOTA black-box UDA methods, DINE \cite{liang2022dine} and BiMem \cite{Zhang_2023_ICCV}, as baselines. 
%We integrate SeCo with them in the same way as in source-free UDA.

In all tasks, we employ CBST \cite{zou2018unsupervised} for generating pixel-level pseudo-labels, and then refine them by SeCo. Subsequently, we utilize the SAM \cite{kirillov2023segany} with Vision Transformer-H (ViT-H) \cite{dosovitskiy2020image} to generate connectivities.
We refrain from using SAM to refine pseudo-labels for test data to avoid introducing extra inference overhead. The automatic mask generation process in SAM adheres to the official parameter settings. In Algorithm~\ref{InSAM}, the enlargement factor for the bounding box area is set to $1.5$.
The connectivity classifier is trained only for 5000 iterations in an early learning way for all tasks.
The noise threshold ($\tau_{ns}$) and correction threshold ($\tau_{cr}$) are configured at 0.60 and 0.95, respectively.

\begin{table*}[htbp]
  \centering
\resizebox{0.975\textwidth}{!}{
    \begin{tabular}{c|cccccccccccccccc|c}
    \toprule
          & \begin{sideways}Road\end{sideways} & \begin{sideways}S.walk\end{sideways} & \begin{sideways}Build.\end{sideways} & \begin{sideways}Wall\end{sideways} & \begin{sideways}Fence\end{sideways} & \begin{sideways}Pole\end{sideways} & \begin{sideways}Tr.Light\end{sideways} & \begin{sideways}sign\end{sideways} & \begin{sideways}Veget.\end{sideways} & \begin{sideways}Sky\end{sideways} & \begin{sideways}Person\end{sideways} & \begin{sideways}Rider\end{sideways} & \begin{sideways}Car\end{sideways} & \begin{sideways}Bus\end{sideways} & \begin{sideways}M.bike\end{sideways} & \begin{sideways}Bike\end{sideways} & mIoU \\
    \midrule
    \multicolumn{18}{c}{\textbf{Unspervised domain adaptation: SYNTHIA $\rightarrow$ Cityscapes}} \\
    \midrule
    AdvEnt \cite{ADVENT_2019_CVPR} $\rm ^{ICCV'19}$  & 87.0  & 44.1  & 79.7  & 9.6   & 0.6   & 24.3  & 4.8   & 7.2   & 80.1  & 83.6  & 56.4  & 23.7  & 72.7  & 32.6  & 12.8  & 33.7  & 40.8  \\
\rowcolor{gray!15}
    AdvEnt + Ours & 87.9  & 47.7  & 82.9  & 20.1  & 1.1   & 38.2  & 29.2  & 28.6  & 86.5  & 85.7  & 64.5  & 29.6  & 84.5  & 44.3  & 39.1  & 47.4  & \textbf{51.1} \rm \textcolor[RGB]{0,155,0}{(+10.3)}  \\
    \midrule
    ProDA \cite{zhang2021prototypical} $\rm ^{CVPR'21}$ & 87.1  & 44.0  & 83.2  & 26.9  & 0.7   & 42.0  & 45.8  & 34.2  & 86.7  & 81.3  & 68.4  & 22.1  & 87.7  & 50.0  & 31.4  & 38.6  & 51.9  \\\rowcolor{gray!15}
    ProDA + Ours & 88.1 & 49.8  & 86.9  & 33.9  & 1.4   & 46.6  & 54.3  & 44.7  & 85.8  & 85.7  & 84.1  & 40.3  & 86.0  & 55.2  & 45.0  & 50.6  & \textbf{58.6} \rm \textcolor[RGB]{0,155,0}{(+6.7)}  \\
    \midrule
    DAFormer \cite{hoyer2022daformer} $\rm ^{CVPR'22}$  & 84.5  & 40.7  & 88.4  & 41.5  & 6.5   & 50.0  & 55.0  & 54.6  & 86.0  & 89.8  & 73.2  & 48.2  & 87.2  & 53.2  & 53.9  & 61.7  & 60.9  \\\rowcolor{gray!15}
    DAFormer+Ours & 88.9  & 49.9  & 90.7  & 46.2  & 7.3   & 55.0  & 63.2  & 57.8  & 87.7  & 92.7  & 76.0  & 51.5  & 89.5  & 61.3  & 59.7  & 64.9  & \textbf{65.1} \rm \textcolor[RGB]{0,155,0}{(+4.2)}  \\
    \midrule
    HRDA  \cite{hoyer2022hrda} $\rm ^{ECCV'22}$ & 85.2  & 47.7  & 88.8  & 49.5  & 4.8   & 57.2  & 65.7  & 60.9  & 85.3  & 92.9  & 79.4  & 52.8  & 89.0  & 64.7  & 63.9  & 64.9  & 65.8  \\\rowcolor{gray!15}
    HRDA+Ours & 90.7  & 50.6  & 89.8  & 51.6  & 8.4   & 59.4  & 66.9  & 64.9  & 89.1  & 95.5  & 81.9  & 58.2  & 91.4  & 66.3  & 65.4  & 66.1  & \textbf{68.5}\rm \textcolor[RGB]{0,155,0}{(+2.3)}  \\
    \midrule
    \multicolumn{18}{c}{\textbf{Source-free domain adaptation:  SYNTHIA $\rightarrow$ Cityscapes} } \\
    \midrule
    HCL  \cite{huang2021model} $\rm ^{NIPS'21}$ & 80.9  & 34.9  & 76.7  & 6.6   & 0.2   & 36.1  & 20.1  & 28.2  & 79.1  & 83.1  & 55.6  & 25.6  & 78.8  & 32.7  & 24.1  & 32.7  & 43.5  \\ \rowcolor{gray!15}
    HCL+Ours & 88.3  & 46.0  & 83.3  & 10.6  & 1.5   & 38.6  & 29.3  & 29.0  & 86.9  & 86.0  & 64.6  & 30.0  & 84.7  & 44.7  & 39.2  & 47.7  & \textbf{50.7} \rm \textcolor[RGB]{0,155,0}{(+7.2)} \\
    \midrule
    DTST  \cite{zhao2023towards} $\rm ^{CVPR'23}$ & 79.4  & 41.4  & 73.9  & 5.9   & 1.5   & 30.6  & 35.3  & 19.8  & 86.0  & 86.0  & 63.8  & 28.6  & 86.3  & 36.6  & 35.2  & 53.2  & 47.7  \\\rowcolor{gray!15}
    DTST+Ours & 88.7  & 48.5  & 87.4  & 23.5  & 2.3   & 39.2  & 30.3  & 31.9  & 91.1  & 86.8  & 64.7  & 33.4  & 88.6  & 45.1  & 43.3  & 57.9  & \textbf{53.9} \rm \textcolor[RGB]{0,155,0}{(+6.2)}  \\
    \midrule
    \multicolumn{18}{c}{\textbf{Black-box domain adaptation:  SYNTHIA $\rightarrow$ Cityscapes}} \\
    \midrule
    DINE \cite{liang2022dine} $\rm ^{CVPR'22}$ & 77.5  & 29.6  & 79.5  & 4.3   & 0.3   & 39.0  & 21.3  & 13.9  & 81.8  & 68.9  & 66.6  & 13.9  & 71.7  & 33.9  & 34.2  & 18.6  & 40.9  \\\rowcolor{gray!15}
    DINE+Ours & 86.7  & 43.9  & 82.1  & 6.8   & 0.0   & 32.5  & 28.3  & 26.7  & 82.1  & 83.9  & 60.0  & 25.1  & 79.1  & 39.8  & 36.5  & 45.8  & \textbf{47.5} \rm \textcolor[RGB]{0,155,0}{(+6.6)}  \\
    \midrule
    BiMem \cite{Zhang_2023_ICCV} $\rm ^{ICCV'23}$ & 78.8  & 30.5  & 80.4  & 5.9   & 0.1   & 39.2  & 21.6  & 15.0  & 84.7  & 74.3  & 66.8  & 14.1  & 73.3  & 36.0  & 32.3  & 21.8  & 42.2  \\\rowcolor{gray!15}
    BiMem+Ours & 84.5  & 43.8  & 79.2  & 8.1   & 0.9   & 39.8  & 25.3  & 25.6  & 85.7  & 85.1  & 63.4  & 29.7  & 82.8  & 40.9  & 35.9  & 44.2  & \textbf{48.4} \rm \textcolor[RGB]{0,155,0}{(+6.2)}  \\
    \bottomrule
    \end{tabular}%\
}
% \setlength{\abovecaptionskip}{0.15 cm}
% \vspace{-.02in}
\caption{Comparison of mIoU scores (\%) for the segmentation performance, where SYNTHIA serves as the source domain.
}
\vspace{-.2in}
  \label{SYN}%
\end{table*}%

\begin{table}[t]
  \centering
\resizebox{0.485\textwidth}{!}{
    \begin{tabular}{c|ccc|c|c}
    \toprule
    \multicolumn{1}{c|}{\multirow{2}[2]{*}{Source\newline{}GTA→}} & \multicolumn{3}{c|}{Compound} & Open  & Average \\
          & Rainy & Snowy & Cloudy & Overcast & C+O \\
    \midrule
    \midrule
    Source Only & 28.7  & 29.1  & 33.1  & 32.5  & 30.9 \\    
    \midrule
    \multicolumn{6}{c}{Unspervised domain adaptation: GTA $\rightarrow$ BDD100k} \\
    \midrule
		ML-BPM \cite{pan2022ml} $\rm ^{ECCV'22}$ & 40.5  & 39.9  & 42.1  & 40.9  & 40.9 \\
		OSC \cite{feng2023open} $\rm ^{NIPS'23}$ & -  & - & -  & -  & 44.0 \\
		\midrule
    PyCDA \cite{Liu_2020_CVPR} $\rm ^{CVPR'20}$  & 33.4  & 32.5  & 36.7  & 37.8  & 35.1 \\  \rowcolor{gray!15}
    PyCDA+Ours & 43.6  & 42.1  & 49.7  & 50.7  & 46.5 \rm \textcolor[RGB]{0,155,0}{(+11.4)} \\  
    \midrule
    ProDA $\rm ^{CVPR'21}$ & 40.3  & 40.6  & 43.2  & 42.5  & 41.7 \\ \rowcolor{gray!15}
    ProDA + Ours & 47.6  & 45.7  & 51.9  & 52.6  &  \textbf{49.5} \rm \textcolor[RGB]{0,155,0}{(+7.8)} \\  
    \midrule
    \multicolumn{6}{c}{Source-free domain adaptation: GTA $\rightarrow$ BDD100k} \\
    \midrule
    SFOCDA \cite{zhao2022source} $\rm ^{TCSVT'22}$ & 35.4  & 33.4  & 41.4  & 41.2  & 37.9 \\  \rowcolor{gray!15}
    SFOCDA+Ours & 41.7  & 42.1  & 44.7  & 47.9  & \textbf{44.1} \rm \textcolor[RGB]{0,155,0}{(+6.2)} \\
    \bottomrule
    \end{tabular}%
}
\vspace{-.01in}
\caption{Comparison of mIoU scores (\%) for the segmentation performance between SeCo (Ours) and other state-of-the-art methods in Open Compoud domain adaptation tasks.
}
\vspace{-.1in}
  \label{BDD}%
\end{table}%

\subsection{Comparison with State-of-The-Art}

The performance of SeCo is shown in Table~\ref{GTA}, Table~\ref{SYN}, and Table~\ref{BDD}.
Overall, experimental results indicate that SeCo can be integrated with various Unsupervised Domain Adaptation (UDA) methods, significantly enhancing their adaptability, including for current state-of-the-art methods. 
Moreover, SeCo exhibits notable improvements for both source-free and black-box adaptation, overcoming limitations with the source domain data.
%Below, we analyze the performance in each transfer task separately.

\noindent \textbf{GTA $\rightarrow$ Cityscapes.} Results on this direction are reported in Table~\ref{GTA}. 
In the UDA setting, the integration of SeCo with AdvEnt \cite{ADVENT_2019_CVPR} leads to a notable performance improvement, achieving a $13.4\%$ increase in mIoU score. 
This shows that combining SeCo with a simple adversarial training method can achieve performance comparable to more complex self-training methods. Combining SeCo with the higher-performing ProDA \cite{zhang2021prototypical} results in a $9.4\%$ increase in mIoU score, surpassing the current optimal performance using the deep-lab model and establishing a new state-of-the-art (SOTA). This highlights SeCo's significant benefits for UDA with CNN architectures.
When integrated with the high-performing Segformer model, SeCo consistently improves DAFormer by 5.3\% in mIoU score and HRDA by 2.3\% in mIoU score.
In source-free UDA, SeCo exhibits stronger advantages. Without source data, SeCo provides robust self-training, elevating the performance of existing state-of-the-art methods, HCL and DTST, by 10.6\% and 8.5\%, respectively. This achieves performance comparable to traditional UDA methods.
In the more stringent black-box adaptation setting, SeCo remains effective. When integrated with two state-of-the-art black-box methods, DINE and BiMem, SeCo refreshes and improves optimal performance by 7.7\% and 8.5\%, respectively, which further underscores the potential of SeCo in data protection scenarios.

\noindent \textbf{SYNTHIA $\rightarrow$ Cityscapes.} Results on this direction are reported in Table~\ref{SYN}. 
The greater domain disparity between SYNTHIA and the target domain, including differences in imaging perspectives and rendering techniques, intensifies the challenges of this task.
In the UDA setting, SeCo continues to exhibit significant performance improvements on both CNN and Transformer-based architectures, maintaining consistent findings with experiments conducted on GTA $\rightarrow$ Cityscapes. Particularly noteworthy is SeCo's ability to enhance the performance by 10.3\% for AdvEnt, a method with initially weaker performance.
The combination of SeCo with ProDA remains effective, refreshing the state-of-the-art methods for CNN structures and approaching the performance level of DAFormer. Additionally, SeCo boosts the mIoU score of DAFormer by 4.2\% and enhances HRDA by 2.3\%.
Under the Source-free and black-box adaptation settings, SeCo surpasses the current SOTA methods, achieving performance improvements of over 6.0\%.

\noindent \textbf{GTA $\rightarrow$ BDD100k.} Results on this direction are reported in Table~\ref{BDD}. 
This task involves complex mixed-weather adaptation, and SeCo consistently maintains stable performance improvements. SeCo enhances the performance of two baseline methods, PyCDA\cite{Liu_2020_CVPR} and ProDA\cite{zhang2021prototypical}, by 11.4\% and 7.8\%, respectively, establishing itself as the SOTA for this benchmark. In the source-free setting, SeCo achieves a 6.2\% improvement over the SOTA method\cite{zhao2022source}, demonstrating sustained and stable performance gains.

\begin{table}[tbp]
  \centering
\resizebox{0.475\textwidth}{!}{
    \begin{tabular}{c|c|ccc|c|c}
		\toprule
    Baselines & Settings & PSA$^{(b)}$  & PSA   & SCC   & mIoU (\%) & $\triangledown$ \\
    \midrule
    \multirow{4}[2]{*}{ProDA \cite{zhang2021prototypical}} & \multirow{4}[2]{*}{UDA} &       &       &       & 53.7  &\\
          &       & \checkmark     &       &       & 57.9 & +4.3 \\
          &       &       & \checkmark     &       & 59.1 & +5.4 \\
          &       &       & \checkmark     & \checkmark    & \textbf{64.1} & +10.4 \\
    \midrule
    \multirow{4}[2]{*}{DAFormer\cite{hoyer2022daformer}} & \multirow{4}[2]{*}{UDA} &       &       &       & 68.2& \\
          &       &    \checkmark   &       &       & 69.7& +1.5 \\
          &       &       & \checkmark     &       & 70.3& +2.1 \\
          &       &       & \checkmark     & \checkmark     & \textbf{73.4} & +4.2 \\
    \midrule
    \multirow{4}[2]{*}{DTST \cite{zhao2023towards}} & \multirow{4}[2]{*}{SF-UDA} &       &       &       & 52.1& \\
          &       & \checkmark     &       &       & 53.9 & +1.8 \\
          &       &       & \checkmark    &       & 55.1& +3.0 \\
          &       &       & \checkmark     & \checkmark     & \textbf{60.5}& +8.4 \\
    \midrule
    \multirow{4}[2]{*}{BiMem \cite{Zhang_2023_ICCV}} & \multirow{4}[2]{*}{BB-UDA} &       &       &       & 48.2& \\
          &       & \checkmark     &       &       & 49.9 & +1.6 \\
          &       &       & \checkmark     &       & 50.4 & +2.2\\
          &       &       & \checkmark     & \checkmark     & \textbf{56.7}& +8.5 \\
    \bottomrule
    \end{tabular}%
}
  \caption{Ablation experiments of SeCo under various UDA settings on GTA $\rightarrow$ Cityscape adaptation task. PSA: Pixel Semantic Aggregation. SCC: Semantic Connectivity Correction. PSA$^{(b)}$ refers to the interaction with SAM using semantic alignment~\cite{chen2023semantic}, as shown in Fig.~\ref{advantage}. SF-UDA: Source-Free UDA. BB-UDA: Black-Box UDA.}
  \vspace{-.1in}
  \label{alb}%
\end{table}%

\begin{figure*}[tbp]
    \begin{center}
    \centering 
    \includegraphics[width=0.975\textwidth, height=0.425\textwidth]{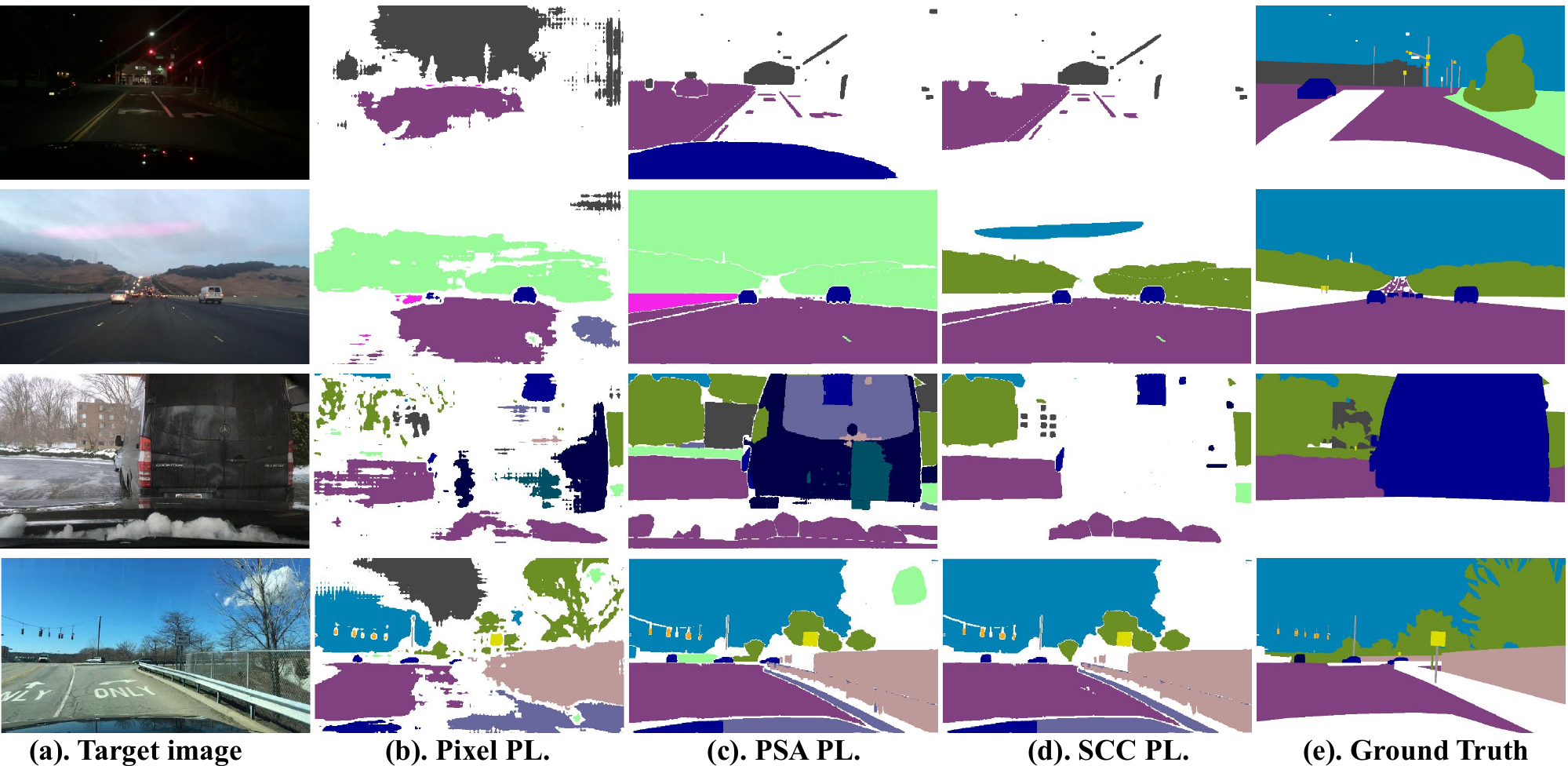}
    \end{center}
    \vspace{-.1in}
%    \captionsetup{font={small}}
    \caption{
Comparison of pseudo-labels generated by original, PSA, and SCC in the open-compound adaptation task  GTA5 $\rightarrow$ BDD100k. White regions in pseudo-label denote filtered areas.
} 
    \label{bdd100k}
\vspace{-.2in}
\end{figure*}

\begin{figure}[tbp]
    \begin{center}
    \centering 
    \includegraphics[width=0.5\textwidth, height=0.225\textwidth]{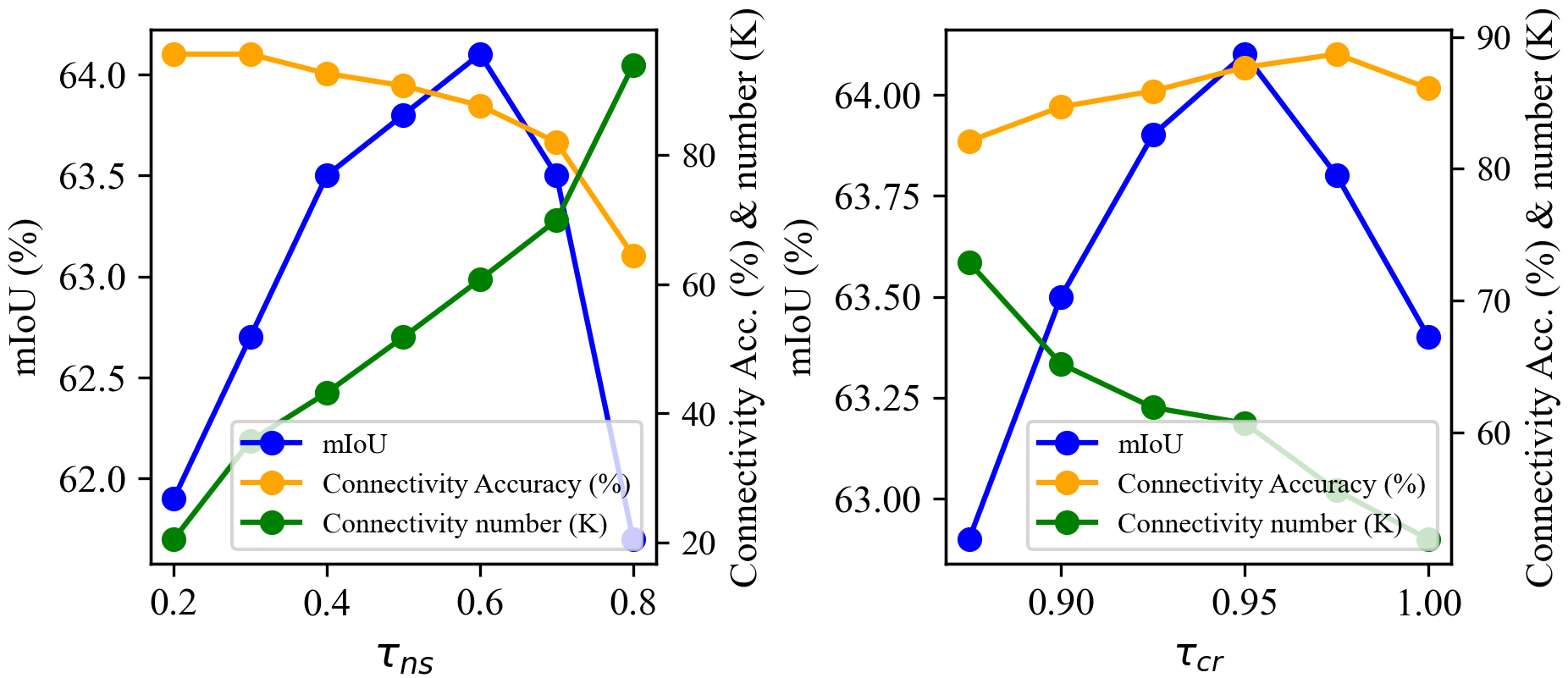}
    \end{center}
%    \setlength{\abovecaptionskip}{10pt} 
    % \setlength{\abovecaptionskip}{-0.1 cm}
%    \captionsetup{font={small}}
    \vspace{-.1in}
    \caption{Evaluation on $\tau_{ns}$ and $\tau_{cr}$ in GTA $\rightarrow$ Cityscapes task using ProDA\cite{zhang2021prototypical} as baseline.
} 
    \label{tau}
\vspace{-.2in}
\end{figure}

\subsection{Analysis}
\noindent \textbf{Ablation Study.} The results of ablation experiments are presented in Table~\ref{alb}. We conducted analyses across different domain adaptation settings as follows. 

\noindent \emph{In UDA}, PSA yields a performance improvement of 5.4\% for ProDA \cite{zhang2021prototypical}, surpassing the gains achieved by PSA$^{(b)}$. Additionally, SCC builds upon PSA, contributing an additional 5.0\% enhancement to ProDA.
A similar trend is observed in the ablation study on DAFormer\cite{hoyer2022daformer}.
These findings suggest that, at the interaction level with SAM, PSA proves more effective than PSA$^{(b)}$; however, interacting solely with SAM is insufficient for achieving substantial self-training performance gains. SCC plays a crucial role in further filtering out noise propagated by SAM, leading to a significant enhancement in UDA performance.

\noindent \emph{In source-Free UDA}, PSA results in a 3.0\% performance improvement for DTST\cite{zhao2023towards}, still outperforming PSA$^{b}$. Due to the substantial initial pseudo-label noise in the source-free setting, PSA aggregates more noisy connections, resulting in a diminished performance gain compared to UDA. SCC, building upon PSA, brings about a 5.4\% performance improvement, reinforcing the notion that SCC can effectively filter and correct propagated pseudo-labels, leading to notable performance gains.

\noindent \emph{In balck-box UDA}, PSA brings about a marginal improvement, with only a 1.2\% performance gain. However, SCC on top of PSA achieves a substantial 7.3\% performance improvement, further confirming the aforementioned conclusions.
These results underscore the importance of correcting noise within connections, especially under more significant domain shifts and weaker initial segmentation results.

\noindent \textbf{Visualization.} 
Fig.~\ref{bdd100k} displays the pseudo-label outputs of PSA and SCC in the GTA5 $\rightarrow$ BDD100K task. In this open compound adaptation task, the model's initial speckled pseudo-labels exhibit considerable noise. It is noticeable that PSA aggregates speckle noise into connected components, concurrently amplifying noisy pseudo-labels. Subsequently, SCC further suppresses and corrects the connected noise from PSA, leading to more structured and lower-noise pseudo-labels. This further validates the motivation behind the design of PSA and SCC.

\noindent \textbf{Hyper-Parameter Sensitivity.} 
Fig.~\ref{tau} illustrates the impact of $\tau_{ns}$ and $\tau_{cr}$ on the final model's adaptability(mIoU), connectivity pseudo-label accuracy (Acc.), and number. 
We set the range of $\tau_{ns}$ from 0.2 to 0.8 to balance excessive connectivity filtering for small values and noise persistence for large values. 
$\tau_{cr}$ is maintained within a confidence threshold range of 0.85 to 0.99 to avoid error correction issues.
It can be observed that within a specific range, the influence of $\tau_{ns}$ and $\tau_{cr}$ on the final model's adaptability is minimal.
Regarding $\tau_{ns}$:
a larger $\tau_{ns}$ retains more connectivity but introduces more noise, leading to a decrease in adaptability.
A smaller $\tau_{ns}$ maintains higher connectivity accuracy, but a lower quantity reduces richness and results in a decrease in mIoU.
Regarding $\tau_{cr}$:
a larger $\tau_{cr}$ corrects some confident connections, improving accuracy and adaptability.
A smaller $\tau_{cr}$ introduces more noise, compromising accuracy and adaptability.
The final $\tau_{ns}$ / $\tau_{cr}$ is set to 0.6 / 0.95.

\section{Conclusion}
In this work,  we propose Semantic Connectivity-driven Pseudo-labeling (SeCo), formulating pseudo-labels at the connectivity level for structured and low-noise semantics. SeCo, comprising Pixel Semantic Aggregation (PSA) and Semantic Connectivity Correction (SCC), efficiently aggregates speckled pseudo-labels into semantic connectivity with SAM. SCC introduces a simple connectivity classification task for locating and correcting connected noise. Experiments demonstrate SeCo's flexibility, significantly improving performance in various cross-domain semantic segmentation tasks, including unsupervised, source-free, and black-box domain adaptation scenarios.
We hope that SeCo can inspire the community to apply SAM to more cross-domain and low-shot semantic segmentation fields.

%%%%%%%%% REFERENCES
{\small
\bibliographystyle{ieee_fullname}
\bibliography{ref}
}

\end{document}